\documentclass[pdflatex,sn-mathphys-num]{sn-jnl}% Math and Physical Sciences Numbered Reference Style
\usepackage{graphicx}%
\usepackage{multirow}%
\usepackage{amsmath,amssymb,amsfonts}%
\usepackage{amsthm}%
\usepackage{mathrsfs}%
\usepackage[title]{appendix}%
\usepackage{xcolor}%
\usepackage{textcomp}%
\usepackage{manyfoot}%
\usepackage{booktabs}%
\usepackage{algorithm}%
\usepackage{algorithmicx}%
\usepackage{algpseudocode}%
\usepackage{listings}%
\usepackage{subcaption}
\usepackage{verbatim}
\theoremstyle{thmstyleone}%
\theoremstyle{thmstyletwo}%

\theoremstyle{thmstylethree}%

\begin{document}

\title[Automatic Rule Extraction for Synthetic Patient Data Generation]{Automatic Extraction of Rules for Generating Synthetic Patient Data From Real-World Population Data Using Glioblastoma as an Example}

%%=============================================================%%
%% GivenName	-> \fnm{Joergen W.}
%% Particle	-> \spfx{van der} -> surname prefix
%% FamilyName	-> \sur{Ploeg}
%% Suffix	-> \sfx{IV}
%% \author*[1,2]{\fnm{Joergen W.} \spfx{van der} \sur{Ploeg} 
%%  \sfx{IV}}\email{iauthor@gmail.com}
%%=============================================================%%

\author[1]{\fnm{Arno} \sur{Appenzeller}}%\email{iauthor@gmail.com}

\author[1]{\fnm{Nick} \sur{Terzer}}%\email{iiauthor@gmail.com}
%\equalcont{These authors contributed equally to this work.}

\author[2]{\fnm{André} \sur{Homeyer}}%\email{iiiauthor@gmail.com}
%\equalcont{These authors contributed equally to this work.}

\author[2]{\fnm{Jan-Philipp} \sur{Redlich}}%\email{iiauthor@gmail.com}

\author[3]{\fnm{Sabine} \sur{Luttmann}}%\email{iiauthor@gmail.com}

\author[4,5]{\fnm{Friedrich} \sur{Feuerhake}}%\email{iiauthor@gmail.com}

%\author[3]{\fnm{Andrea} \sur{Eberle}}%\email{iiauthor@gmail.com}

\author[4]{\fnm{Nadine S.} \sur{Schaadt}}%\email{iiauthor@gmail.com}

\author[6]{\fnm{Timm} \sur{Intemann}}%\email{iiauthor@gmail.com}

\author[7]{\fnm{Sarah} \sur{Teuber-Hanselmann}}%\email{iiauthor@gmail.com}

\author[8]{\fnm{Stefan} \sur{Nikolin}}%\email{iiauthor@gmail.com}

\author[8]{\fnm{Joachim} \sur{Weis}}%\email{iiauthor@gmail.com}

\author[9]{\fnm{Klaus} \sur{Kraywinkel}}%\email{iiauthor@gmail.com}

\author*[1]{\fnm{Pascal} \sur{Birnstill}}\email{pascal.birnstill@iosb.fraunhofer.de}

%\affil*[1]{\orgdiv{Department}, \orgname{Organization}, \orgaddress{\street{Street}, \city{City}, \postcode{100190}, \state{State}, \country{Country}}}

\affil*[1]{\orgname{Fraunhofer Institute of Optronics, System Technologies and Image Exploitation IOSB}, \orgaddress{\street{Fraunhoferstraße 1}, \city{76131 Karlsruhe}, \country{Germany}}}

%\affil[1]{\orgaddress{\street{Fraunhoferstraße 1}, \city{Karlsruhe}, \postcode{76131}, \country{Germany}}}

\affil[2]{\orgname{Fraunhofer Institute for Digital Medicine MEVIS}, \orgaddress{\street{Max-Von-Laue-Straße 2}, \city{28359 Bremen}, \country{Germany}}}

%\affil[3]{\orgdiv{Department}, \orgname{Organization}, \orgaddress{\street{Street}, \city{City}, \postcode{610101}, \state{State}, \country{Country}}}

\affil[3]{\orgname{Bremen Cancer Registry, Leibniz Institute for Prevention Research and Epidemiology - BIPS}, \orgaddress{\street{Achterstraße 30}, \city{28359 Bremen}, \country{Germany}}}

\affil[4]{\orgname{Institute for Pathology, Hannover Medical School}, \orgaddress{\street{Carl-Neuberg-Straße 1}, \city{30615 Hannover}, \country{Germany}}}

\affil[5]{\orgname{Institute of Neuropathology, Medical Center - University of Freiburg}, \orgaddress{\street{Breisacher Straße 64}, \city{79106 Freiburg}, \country{Germany}}}

\affil[6]{\orgname{Research Data Infrastructures and Data Science, Leibniz Institute for Prevention Research and Epidemiology - BIPS}, \orgaddress{\street{Achterstraße 30}, \city{28359 Bremen}, \country{Germany}}}

\affil[7]{\orgname{Department of Neuropathology, Center for Pathology, Klinikum Bremen-Mitte}, \orgaddress{\street{St.-Jürgen-Straße 1}, \city{28205 Bremen}, \country{Germany}}}

\affil[8]{\orgname{Institute of Neuropathology, RWTH Aachen University Hospital}, \orgaddress{\street{Pauwelsstraße 1}, \city{52074 Aachen}, \country{Germany}}}

\affil[9]{\orgname{Robert Koch Institute, Center for Cancer Registry Data}, \orgaddress{\street{Nordufer 20}, \city{13353 Berlin}, \country{Germany}}}

%%==================================%%
%% Sample for unstructured abstract %%
%%==================================%%

\abstract{The generation of synthetic data is a promising technology to make medical data available for secondary use in a privacy-compliant manner.
A popular method for creating realistic patient data is the rule-based \textit{Synthea} data generator.
Synthea generates data based on rules describing the lifetime of a synthetic patient. 
These rules typically express the probability of a condition occurring, such as a disease, depending on factors like age.
Since they only contain statistical information, rules usually have no specific data protection requirements.
However, creating meaningful rules can be a very complex process that requires expert knowledge and realistic sample data.
In this paper, we introduce and evaluate an approach to automatically generate Synthea rules based on statistics from tabular data, which we extracted from cancer reports.
As an example use case, we created a Synthea module for glioblastoma from a real-world dataset and used it to generate a synthetic dataset.
Compared to the original dataset, the synthetic data reproduced known disease courses and mostly retained the statistical properties.
%However, we also observed statistical deviations, which we partly attribute to the high degree of automation of our Synthea rule generation approach.
%Our comparison with the original dataset shows that the synthetic data fits known disease courses and retains some of the statistical properties of the original data.
%In our case, limitations of the Synthea rule modeling capabilities led to more favorable clinical outcomes compared to real-world data.
%We discuss the statistical properties for which we observed deviations in the synthetic data.
%In part we attribute these to the high degree of automation of our Synthea rule generation approach.
Overall, synthetic patient data holds great potential for privacy-preserving research.
The data can be used to formulate hypotheses and to develop prototypes, but medical interpretation should consider the specific limitations as with any currently available approach.
%conclusions should be drawn with caution.
}


%%\pacs[JEL Classification]{D8, H51}

%%\pacs[MSC Classification]{35A01, 65L10, 65L12, 65L20, 65L70}

\maketitle

\section{Background}\label{sec:intro}

The ongoing digitization in the healthcare sector has led to a growing availability of digital medical data.
This data has the potential to improve medical care for patients and research.
Recent advances in the fields of artificial intelligence (AI) have led to ever-increasing demand for medical real-world data from industry and academic research.
%Recently, the demand from industry and researchers has become increasingly strong.
%To collect even more data, networked medical devices can serve as a source of digital medical data for patients.
Medical data is increasingly collected in a patient's personal electronic health record in a standardized way which makes it an attractive resource for research.
%, where it could potentially be used for research.
This has led to regulatory initiatives intended to make medical data widely available while protecting individual privacy.
One example is the "Gesundheitsdaten-Nutzungsgesetz"\footnote{\url{https://www.recht.bund.de/bgbl/1/2024/102/VO.html} (German; last accessed 2025/07/18)} (GDNG; in English: Health Data Usage Act) in Germany.
%This legislation makes the medical data of patients, collected in their personal health record (elektronische Patientenakte, ePA), available for research in a pseudonymized form unless the affected person opts out.
Under this legislation, medical data from patients’ personal health records (elektronische Patientenakte, ePA) may be made available for secondary use in research in pseudonymized form, unless the individual opts out.
While health records are an invaluable data source for research, there are serious privacy concerns regarding the secondary usage of the data~\cite{https://doi.org/10.1111/joim.12119}.
Numerous examples demonstrate that data deemed anonymized can be re-identified, raising concerns that current standards are insufficient to protect the privacy of individual patients~\cite{Sweeney2002}.
Recently, methods for synthetic data generation have been increasingly investigated to enable the exchange of personal health data in a privacy-preserving manner.
%To preserve privacy while sharing personal data, there exists a multitude of so-called privacy-preserving technologies.
%Recently, the concept of synthetically generated data has been increasingly investigated in this field.
The idea is to generate a representative dataset based on the statistical properties of the real dataset, but without allowing re-identification.
Different approaches exist to generate such data.

A common rule-based approach is the \textit{Synthea} data generator~\cite{syntheaRef}.
Synthea uses various types of rules to create synthetic and realistic health records, e.g. following the FHIR format~\cite{fhirStandard}.
%A typical rule for Synthea may consider the age of a patient and provide a probability for the occurrence of the next state, which can be a health condition or a medical encounter.
A typical rule for Synthea may consider the age of a patient and provide a probability for the occurrence of, for example, a health condition or a primary care encounter.
Complex rules can be written to define a realistic health record for a certain condition.
However, there is only a limited set of predefined modules providing realistic scenarios, and they do not adequately cover the entire population or particularly rare diseases.
The complexity of the modeling process contributes to these limitations.
For example, the rule set for allergic rhinitis already contains 300 lines of code in the designated JSON format.

\begin{figure}
	\centering
	\includegraphics[width=\linewidth]{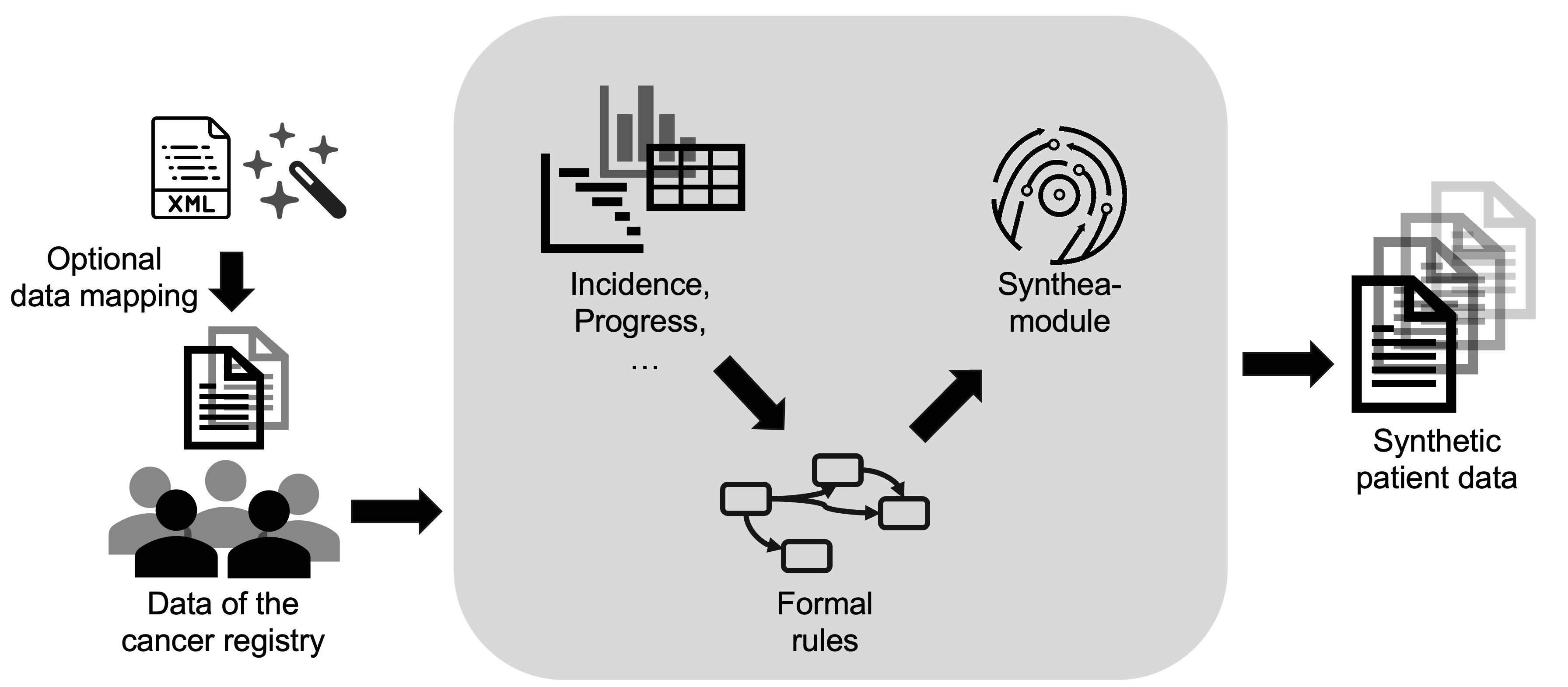}
	\caption{Workflow for automatic Synthea rule generation}
	\label{fig:automatic_rule_generation_flow}
\end{figure}

As an alternative to manual rule creation, this paper introduces a workflow for automatically generating rules for the Synthea generator based on population information (cf.~Figure~\ref{fig:automatic_rule_generation_flow}).
We demonstrate this workflow based on 
%disease incidence data from the German Center for Cancer Registry Data (Zentrum für Krebsregisterdaten, ZfKD) and also based on
case reports from German cancer registries on glioblastoma cases, which were represented in the German oBDS cancer report format (German nationwide oncological basis dataset; from German: onkologischer Basisdatensatz)\footnote{\url{https://www.basisdatensatz.de/basisdatensatz} (German, last accessed 2025/07/21)},  previously also referred to as the ADT/GEKID format\footnote{Arbeitsgemeinschaft Deutscher Tumorzentren e.V. (ADT) und Gesellschaft der epidemiologischen Krebsregister in Deutschland e.V. (GEKID)}.
%To facilitate the manual process of rule generation, this paper introduces a process to automatically generate rules for the Synthea data generator based on statistical population information, such as disease incidence.
%We demonstrate our approach using data from three German cancer registries and also present an integration with the German oBDS cancer report format (German nationwide oncological basis dataset; from German: onkologischer Basisdatensatz)\footnote{\url{https://www.basisdatensatz.de/basisdatensatz} (German, last accessed 2025/07/21)} used by the cancer registries in Germany.
%The oBDS format was previously also referred to as ADT/GEKID format\footnote{Arbeitsgemeinschaft Deutscher Tumorzentren e.V. (ADT) und Gesellschaft der epidemiologischen Krebsregister in Deutschland e.V. (GEKID)}.
%:\\ \url{https://www.basisdatensatz.de} (German, last accessed 2025/07/21)}, which is used by the cancer registries in Germany.
%Data used for Synthea rule extraction were collected during the CanConnect project\footnote{\url{https://www.canconnect-projekt.de/en.html}}, where brain tumor cancer reports from three regional German cancer registries were combined and curated.
Data used for Synthea rule extraction were collected during the CanConnect project, funded by the German Federal Ministry of Health, where data on glioblastoma cases provided by three German cancer registries were combined.
%where brain tumor cancer reports from three regional German cancer registries were combined and curated.
This data provides a unique dataset for glioblastoma, a relatively rare malignancy compared to more common types such as breast and lung cancer.
Although glioblastomas are the most common type of brain tumor, they account for only about 2\% of all cancer cases in Germany.
%In the following, we refer to this dataset as the \textit{CanConnect dataset}.

Our approach is intended to serve as a basis for automatic rule generation from structured population statistics and to facilitate the creation of synthetic data based on a specific set of rules.
%To be more specific, statistical information is extracted from the CanConnect dataset and used to generate Synthea rules.
\begin{comment}
This approach also ensures that no sensitive input data is leaked, which is a risk in machine learning-based approaches~\cite{Stadler2021}.
Such methods could also be directly applied to networked medical devices.
For example, the technology could be implemented on a device to create a rule set for real data collected by the device.
This would allow for the direct creation of a realistic research cohort on-device, which can be used without any privacy implications as no real data needs to leave the device.
\end{comment}

The contributions of this paper are the following:
\begin{itemize}
    %\item A workflow for the automatic extraction of Synthea rules from individual cancerregistry reports
    %\item An implementation of this workflow for data in the oBDS cancer report format
    \item A description and open-source implementation of a workflow for the automatic extraction of Synthea rules from individual cancer registry reports in the oBDS cancer report format
    %\item Workflows and their implementations for automatic generation of Synthea rules from incidence data in CSV format and from individual cancer registry reports in the oBDS cancer report format
    %provided by the CanConnect glioblastoma project
    %\item Implementations of this workflow for incidence data of the German Centre for Cancer Registry Data (ZfKD data, cf.~\ref{subsec:dataset}), and for data in the oBDS cancer report format, whereby we applied the latter implementation to the CanConnect project's glioblastoma dataset provided by the participating cancer registries
   % \item A comparison of a generated synthetic dataset  using the presented approach with the CanConnect dataset used to generate the Synthea rules
    %\item An application to the CanConnect project's glioblastoma dataset provided by the participating cancer registries, which delivers a Synthea module for glioblastoma
    %\item A comparison of a synthetic dataset generated with this glioblastoma Synthea module with the original CanConnect dataset
    \item An exemplary application of this workflow to the CanConnect dataset, resulting in the creation of a Synthea module for glioblastomas and a comparison of a synthetic dataset generated using this module with the original dataset
\end{itemize}
%Note that we also created a generic Synthea module for all types of cancer from incidence data of the German Centre for Cancer Registry Data (ZfKD data, cf.~\ref{subsec:dataset}).
%However, we did not have any real datasets available for evaluation.
%The software artifacts developed for this work (oBDS parser and glioblastoma Synthea module) are available as open source code.\footnote{\url{https://gitlab.cc-asp.fraunhofer.de/canconnect/synthea-canconnect-generator}}

%The remainder of this work is structured as follows:
%Section \ref{sec:relwork} provides an overview of related work on the topic.
%Section \ref{sec:prelim} introduces Synthea, the required data formats, the datasets used in our work, and our Synthea rule creation workflow.
%Based on this approach and our dataset, we conduct an evaluation in Section \ref{sec:eval}.
%We discuss our observations in Section~\ref{sec:discussion} and conclude the paper in Section~\ref{sec:conclusion}.

\subsection{Related Work}\label{sec:relwork}

Among the numerous published approaches to generating synthetic medical data, the following works are particularly relevant.

%Data-driven approach for creating synthetic electronic medical records
%https://link.springer.com/article/10.1186/1472-6947-10-59
Buczak et al. described an approach where real world health records are used to identify patterns common to specific health conditions~\cite{Buczak2010}.
Through this, a series of steps that represent typical treatments or common conditions can be identified.
%Another use case is tracking the progress of a disease.
These patterns can then be used to create synthetic patient data.
%One example in the paper is the outbreak of an illness.
%The characteristics of the outbreak define the input population for the synthetic data.
%By applying the identified patterns, a synthetic dataset was created.
While this approach appears to be similar to the rule-based approach of Synthea, a key difference is that patterns resemble entire pathways whereas Synthea rules describe conditions for state changes within a patient's pathway.  
The latter approach appears to be more suitable for modeling and reviewing medical correlations that influence patients' pathways during synthetic data generation.

%Differentially Private Medical Texts Generation Using Generative Neural Networks
%https://dl.acm.org/doi/abs/10.1145/3469035
Al Aziz et al.~investigated the generation of medical texts by combining neural networks to generate the data and differential privacy to preserve the privacy of the input data~\cite{10.1145/3469035}.
A transformer language model was applied to the input data from medical records to generate new data.
Since the input dataset could contain sensitive information that might be preserved in the model, Gaussian noise was added to the weights of the model to maintain differential privacy guarantees.
The evaluation shows that the generated data achieves good scores in the utility metrics chosen by the authors and is almost equally suitable for training machine learning models as the original dataset.
%However, the authors note that an expert evaluation is needed.
%In contrast to the work presented in this paper, the article by Al Aziz et al.~demonstrates that generative models provide an effective way to generate synthetic data.
%However, fine-tuning the model only works by altering the network, and medical relationships are harder to realize than when using rule-based models like Synthea.

%Synthetic data generation for tabular health records: A systematic review
%https://www.sciencedirect.com/science/article/pii/S0925231222004349
Hernandez et al.~\cite{HERNANDEZ202228} conducted a systematic review of synthetic data generation for tabular health records.
While they examined different machine learning models, particularly Generative Adversarial Models (GAN), as well as methods like Synthea, the authors concluded that there is no universal method for synthetic data generation.
The authors also mention that GANs are often very use case-specific and need to be customized depending on the domain.
While this paper primarily examines machine learning-based methods for synthetic data generation, the findings emphasize the importance of understandable rule-based generators like Synthea.
%This also highlights the significance of automatic rule generators that can combine the simplicity of a GAN-based approach with the flexibility of a rule-based method.

%Synthea™ Novel coronavirus (COVID-19) model and synthetic data set
%https://www.sciencedirect.com/science/article/pii/S2666521220300077
Walonoski and others demonstrated an exemplary use of Synthea to model rules for a synthetic dataset for COVID-19~\cite{WALONOSKI2020100007}.
For this, findings from three different studies were used to identify the required characteristics, such as outcomes, symptoms, and complications.
As a result, a cohort of 124,000 patients was created.
%The authors showcase the quality of the data through various visualizations.
This model illustrates that even with limited data, synthetic data can be generated, which faithfully models the original dataset.
However, the rule creation was a manual process and is not described in detail.
%This leads to the assumption that this process is highly dependent on expert knowledge.

In summary, the previous related work highlights the importance of generating synthetic data.
%significance of the general topic of synthetic data generation is evidenced in the related work.
However, while there are many machine learning-based approaches, studies focusing on rule-based methods and their automatization are lacking.
In contrast to the ``black-box'' nature of many types of machine learning models, the advantages of rule-based approaches are better explainability, greater transparency, and easier to achieve privacy.
%Medical relationships can be assumed to be easier to observe in rules.
%Thus, customizing rules seems accessible to medical experts in contrast to analyzing and fine-tuning of machine learning models.
As many real-world medical decision processes follow a rule-based pattern, this approach is expected to be more familiar for medical experts than less explainable ``black-box'' approaches based on machine learning.
This improves perspectives for adaptation and acceptance in clinical practice. 

\section{Methods}\label{sec:prelim}

%This section describes the \textit{Synthea} data generator in detail as well as the data sets that were used for this paper.

\subsection{Synthea}
\label{subsec:synthea}
%More details about Synthea
%https://academic.oup.com/jamia/article/25/3/230/4098271?login=false
The Synthea data generator was created by Walonoski et al.~in 2018~\cite{walonoski2018synthea} and aims to generate synthetic medical data, i.\ e.\ health records, using only publicly available information.
It is based on rules defined through medical workflows and observations.
Synthea starts with demographic information for a geographic region. Using these demographics, Synthea randomly creates individuals with realistic age, gender, etc.\ for that region.
Synthea simulates the entire life of a patient from birth until their death or the current day, resulting in health records that include diverse observations from a whole lifetime and not just from a specific disease.
%Synthetic patient records generated with Synthea can be exported in multiple formats including FHIR~\cite{fhirStandard} and CSV.
%, e.g. HL7 FHIR\footnote{\url{https://hl7.org/fhir/}, (last accessed 2025/07/21)} or CSV. 

In order to train machine learning (ML) models such as neural networks to generate synthetic data, the training process must receive real data as input.
%Subsequently, it cannot be excluded that a reconstruction of training data from the ML model is possible.
Consequently, it cannot be ruled out that it is possible to reconstruct information about the training data from the ML model.
Privacy-enhancing technologies would have to be used as a preventative measure during the training process.
%In contrast, the concept of Synthea ensures that original data and synthetic data remain completely separate.
In contrast, the concept of Synthea ensures that conclusions from the synthetic data cannot be drawn.
This is achieved by using only aggregated information derived from individual health records.
This aggregated information is used to create rules that define the potential states patients can reach during their lifetime for particular disease progressions.
These rules are defined in modules that are applied throughout a patient’s simulated life.
Each module typically starts with an initial state, representing the beginning of a medical condition, a health event, or a routine checkup.
From there, transitions between states occur based on conditional transition probabilities obtained from the aggregated information, logical conditions, or time-based progressions.
To ensure that no sensitive information can be recovered from a Synthea module, it must be ensured that the data fed into the rule generation workflow meets appropriate privacy metrics.
We elaborate on this issue in the following section.

Synthea states and transitions define patient pathways, simulating disease progression, treatments, and primary care encounters.
Modules in Synthea can model various aspects of healthcare, such as chronic diseases (e.g., diabetes, hypertension), acute conditions (e.g., fractures, infections), and preventive care (e.g., vaccinations, screenings).
Each module is structured using Synthea's Generic Module Framework, where states can include all sorts of actions such as adding conditions, prescribing medications, performing procedures, or recording symptoms.
These actions are driven by real-world clinical guidelines as well as relative frequencies and event time distributions for events related to disease progression derived from public health data sources.
\begin{figure}
    \centering
    \includegraphics[width=\linewidth]{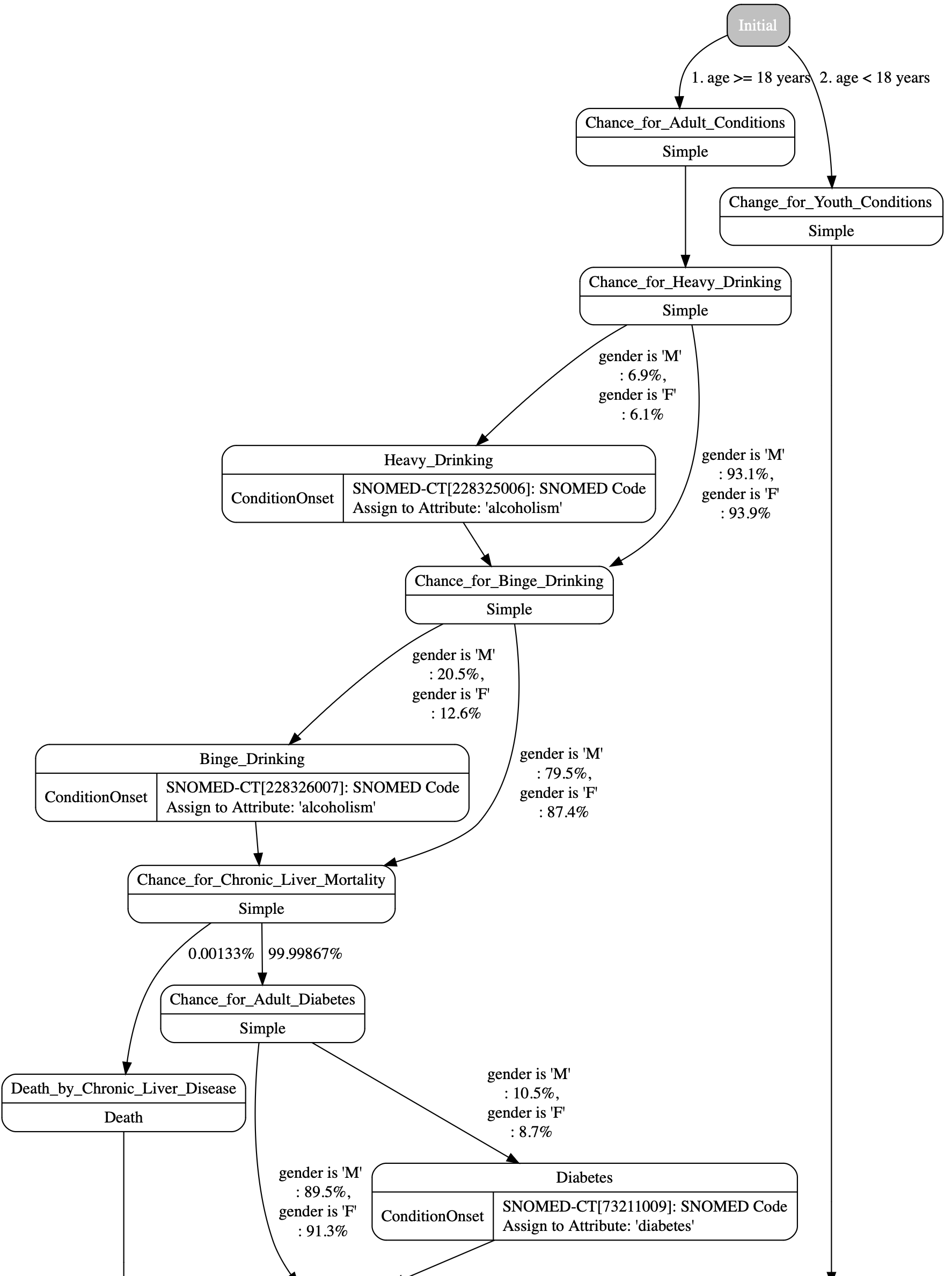}
    \caption{Excerpt of a \textit{Synthea} module for alcohol abuse visualized in the \textit{Synthea} Module Builder (boxes stand for states, arrows for state transitions depending on rules; Simple states are e.g. used to chain conditional transitions, ConditionOnset states mean that the patient acquires the given condition)}
    \label{fig:enter-label}
\end{figure}
Figure \ref{fig:enter-label} shows an example module.

Additionally, Synthea enables the customization and extension of its modules, allowing researchers to define new disease modules, intervention strategies, or demographic characteristics.
By adjusting transition probabilities, modifying state sequences, or integrating new observational data, users can tailor simulations to specific research questions or healthcare scenarios.
Synthetic data can be generated for multiple output formats, including FHIR, CSV, and OMOP, facilitating integration with health research pipelines, machine learning applications, and healthcare system simulations.
%standard modules
%population data https://github.com/synthetichealth/synthea/wiki/Default-Demographic-Data

\subsection{Dataset}
\label{subsec:dataset}

The  dataset used in this study was collected during the CanConnect project.
%, funded by the German Federal Ministry of Health.
%This dataset contains real-world report data from 1441 glioblastoma patients (with histological findings), provided by three cancer registries from different German states.
This dataset contains real-world report data on 1,441 glioblastoma patients, as reported by the neuropathology departments of three hospitals to three different cancer registries.
Glioblastoma is the most common malignant form of brain cancer, though it is relatively rare compared to breast or lung cancer.
%The second dataset is the CanConnect glioblastoma dataset with individual cancer registry data from glioblastoma patients with histological findings compiled by the pathologists participating in the CanConnect project.
%The second dataset consists of data provided to the CanConnect project by three German cancer registries, the Bremen Cancer Registry, the Clinical Cancer Registry Lower Saxony, and the State Cancer Registry of North-Rhine Westphalia.
%In the cancer registries, the individual oBDS reports are checked for plausibility, processed, and summarized into a complete data set containing the best information on each tumor.
The CanConnect dataset is structured as a relational database.
%The data contains different types of glioblastoma including particularly few cases for the tumor localizations $C71.5$, $C71.6$, $C71.7$, and $C72.0$, which are actually rarer localizations of glioblastomas.
The data contains different types of glioblastoma including common localizations (e.g., $C71.1$-$C71.4$, $C71.8$), reflecting different supratentorial tumor sites), few cases for the less common tumor localizations ($C71.5$, $C71.6$, $C71.7$, $C72.0$, where $C72.0$ is actually a tumor in the spinal cord), and the generic code $C71.9$ for CNS, NOS.
Glioblastomas have an extremely poor prognosis, as can be seen from the survival times after diagnosis (cf.~Section~\ref{subsec:survival-time}). 
% with particularly few cases for the tumor types $C71.5$, $C71.6$, $C71.7$, and $C72.0$ (35 cases for these four types combined and correspondingly even fewer cases where it is known how long the patients survived after these diagnoses).
Different reports during a patient's course of disease have been summarized in one data record containing all therapies and treatments.
%For each case it contains information including diagnoses, surgical interventions, and systematic therapies. 
In accordance with the project's data protection concept, dates are not stored exactly in the data set, but only relatively as days since diagnosis.
Place of residence data is only stored at the state level (but was not used for the work presented in this article).
Age data is stored in 5-year groups, while the time of initial diagnosis is given as a year.
For the work in this paper we particularly used dates of diagnosis, surgical procedures, systemic therapies, radiotherapies, and date of death.
If available, these data are given in days after diagnosis, with systemic therapies and radiotherapies having a start and end date.
The CanConnect dataset was used to create the Synthea module presented in the next section.

%When creating Synthea rules, a privacy risk can arise in particular when case groups with few cases are processed.
%This can lead, for example, to mean values and standard deviations being represented in the Synthea module that are greatly influenced by outliers.
%This can enable membership inference attacks on a Synthea module, for example, if a case with a particularly long survival time after diagnosis was included in the input data.
%To minimize this risk, we identified rare case groups and set k-anonymity to $k=5$ as requirement.
%We found that the most critical case groups can be characterized as gender $\times$ primary tumor localization $\times$ survival time after diagnosis in days.
%It turns out that we achieve $k=5$ for these case groups by removing the rare tumor localizations described above from the dataset.
%For all other tumor localizations we verified that we also achieve $k=5$ for outliers in survival times after diagnosis.
%Due to these privacy characteristics of the CanConnect dataset (without the rare tumor localizations), we argue that the information encoded into Synthea rules cannot be used to draw conclusions about specific individual cases of the CanConnect dataset.
When creating Synthea rules, a privacy risk can arise in particular when case groups with few cases are processed.
To minimize this risk, we identified rare case groups and set k-anonymity to $k=10$ as requirement.
We found that the most critical case groups can be characterized as gender $\times$ primary tumor localization $\times$ deceased.
It turns out that we achieve $k=14$ for these case groups by removing the rare tumor localizations described above from the dataset.
Due to these privacy characteristics of the CanConnect dataset (without the rare tumor localizations) and the data minimization measures already implemented in the CanConnect data set, we argue that the information encoded into Synthea rules cannot be used to draw conclusions about specific individual cases of the CanConnect dataset.

In order to obtain a generic and reusable implementation of Synthea module generation we use the oBDS format for input data, which is the standard for exchanging oncological data between healthcare institutions and cancer registries in Germany and enables structured and interoperable reporting of cancer cases.
Accordingly, the relational database was mapped to the oBDS format.

\subsection{Data mapping}
\label{subsec:data-mapping}
%The previously described proof of concept was based on individual datasets and did not consider a multi-table database structure, such as the cancer registry data provided in the CanConnect project.
%In contrast to the ZfKD dataset used for the previously described proof of concept, 
The dataset provided by the cancer registries in the CanConnect project has a multi-table database structure.
To align this data with our tool chain for generating Synthea rules, a data mapping process was necessary to convert the data into the oBDS format.
%This mapping was performed based on the reference datasets of the oBDS format.
First, fictitious exact data for the date of birth and date of diagnosis were generated.
As described before, the age at diagnosis was available by age group (cf.~\ref{subsec:dataset}).
An exact fictitious date of birth was generated for each patient.
To do this, a random value was drawn from within the respective age interval (the fictitious exact age at diagnosis).
The fictitious exact date of birth was then calculated by subtracting the fictitious exact age at diagnosis from the date of diagnosis. This was also drawn randomly from the given year of diagnosis.
Individual oBDS reports were then generated for all other data, i.~e.\, medical encounters, surgeries, systematic therapies (start and end), radiotherapies (start and end), and date of death.
Note that the CanConnect dataset is not free of missings.
If, for example, several months after diagnosis there is still no report with a therapy or date of death, then it can be assumed that the data for the case is incomplete.
However, we did not sort out such cases.
As a consequence, with our approach described in the next section, we also generate incomplete synthetic cases with similar frequency.
%After mapping, individual patient-level records were extracted from the provided tables and structured in accordance with cancer registry reports.
A challenge emerged regarding the matching of OPS\footnote{\url{https://www.bfarm.de/EN/Code-systems/Classifications/OPS-ICHI/OPS/_node.html}, (last accessed 2025/07/21)} (Operation and Procedure Codes) and systematic therapies.
In real-world scenarios, a single cancer registry report can include multiple therapies and even different therapy types for the reporting time point.
If there were multiple therapies with the same date, a separate report in the oBDS format was generated for each therapy.
%As further explained in the following paragraph, we simplify by only processing the first therapy per therapy type.
The mapped data therefore contains all medical encounters, all surgical procedures with OPS codes, all systemic therapies, and all radiotherapies in the order in which they appear in the database form of the CanConnect dataset.
As the CanConnect dataset does not include exact dates, but only days after diagnosis, the reporting dates of the generated records were calculated based on tumor diagnosis dates.

\subsection{oBDS Parser}
\label{subsec:obds-parser}
The generated records in the oBDS format were processed by a Module Parser to create a Synthea module for the glioblastoma use case. 
The Module Parser analyzes therapy options and combinations based on their relative frequency in order to model possible treatment pathways using Synthea states with conditional transition probabilities.
%The Module Parser analyzes therapy options and combinations to model possible therapy pathways in Synthea states.
The CanConnect dataset used for this publication shows a huge variety of pathways.
%Partly this is due to only minimal data cleansing of the data set.
%We justify applying no further data cleansing by our goal of investigating the automatability of generating Synthea rules and synthetic data.
%To manage this complexity and ensure computational feasibility, in the first version of the Module Parser a restriction was applied by only considering the first therapy per therapy type and ignoring further therapies of the same type (cf.~Section~\ref{subsec:data-mapping}).
In order to deal with this complexity and ensure computational feasibility, we simplified matters for the present study letting the Module Parser only consider the first therapy per therapy type and ignoring further therapies of the same type. % (cf.~Section~\ref{subsec:data-mapping}).
In terms of possible courses of treatment to be generated, this means that an individual may have one or no surgical procedure, one or no systemic therapy, and one or no radiotherapy.
Furthermore, while we do not consider combinations of OPS codes for surgeries, we for example consider multiple substances at the same time in systemic therapies.
%While this simplification reduces the expressiveness of the generated courses of treatment, it ensures computational feasibility.

The general idea behind our approach to analyzing therapy options is to determine, for each condition, the conditional transition probabilities to subsequent conditions and the typical time intervals between them.
%The general idea behind our approach to analyzing therapy options is that we determine for each condition the conditional transition probability of which next condition will follow, and how much time typically elapses between these conditions.
%The general procedure here is to determine relative frequencies and transition probabilities for all pairs of entries that follow each other in the reports.
To do this, we proceeded as follows:
\begin{enumerate}
    \item Read all cases from the oBDS reports and create a table with the columns gender, date of birth, primary tumor localization, and date of diagnosis.
    \item Add the remaining data to each case (surgical procedures, systematic therapies, radiotherapies, and date of death), if available.
    \item Determine the conditional transition probabilities to glioblastoma diagnoses as relative frequencies of tumor localizations in the CanConnect cases depending on gender.
    \item Estimate gender-specific Gaussian distribution of age at diagnosis based on estimated mean and standard deviation.
    \item Prepare the table for extracting Synthea states:
    \begin{itemize}
      \item Add date of birth as start.
      \item Add date of last report or date of death as end. 
      \item For systematic therapies and radiotherapies, separate start and end dates.
      \item Interim result: a table in which all elements are listed in chronological order
    \end{itemize}
    \item Identify Synthea states (gender-specific):
    \begin{itemize}
      \item Determine which different surgical procedures, systematic therapies, and radiotherapies appear in the table.
      \item For all pairs of elements that follow each other, determine how often this transition occurs and how much time typically passes between them (minimum and maximum, averaged over all occurrences).
      \item For transistions to a date of death: estimate exponential distributions (supported by Synthea) using the estimated mean.
      \item Interim result: a table with conditional transition probabilities between states and a table with minimum and maximum time intervals between states; uniform distributions are assumed for time invervals between therapies
    \end{itemize}
    \item Create Synthea states: All possible courses of events assembled into a Synthea module (a type of state machine for simulating glioblastoma disease in the life of a synthetic patient) in JSON format:
    \begin{itemize}
    	\item Begin state
    	\item Transitions to diagnoses
    	\item Transitions to therapies (different surgical procedures, systemic therapies, and radiotherapies)
    	\item Up to two additional transitions to a surgical procedure, systemic therapy, or radiotherapy
		\item Transitions to the date of death or directly to the end state
    \end{itemize}
    %\item We determine normal distributions (mean and standard deviation) for the age at diagnosis.
    %\item We determine exponential distributions (mean) for survival time after diagnosis in days.
\end{enumerate}
Note that the simplification described before is applied in step $2$.
It should also be noted that loss to follow-up naturally occurs in such observational datasets.
That means, the estimated exponential distributions do not represent the actual survival times, but serve the purpose of reflecting the data that contains loss to follow-up.
Actual survival times should then be estimated using appropriate methods of event time analysis (such as Kaplan-Meier estimators) for both the original and the synthetic dataset.
%in step $2$ we simplified the data by taking a maximum of one surgery, one systematic therapy, and one radiation therapy (the first of each) from each case.

In terms of Synthea states and rules, branches in patients' pathways, such as a diagnosis of a specific glioblastoma localization, were depicted as \textit{distributed\_transitions} with conditional transistion probabilities derived from the CanConnect dataset.
The creation of dates such as the date of diagnosis or the date of death was modeled using \textit{delay states} containing distributions and their parameters.
A delay state ensures that a condition, such as a diagnosis, is not assigned to a patient before a minimum period of time has elapsed.
As described before, for delays before diagnosis Gaussian distributions were used, for delays before date of death (survival times), exponential distributions were used, and finally for delays between therapies, uniform distributions were used.
As an example, a diagnosis is then assigned to a synthetic patient by a Synthea rule with a transition probability and a randomly generated delay.
The resulting Synthea module for glioblastoma comprises 418 individual states.
Due to this complexity, it cannot be shown in a meaningful way in this publication.

The implemented software artifacts, i.\ e.\ oBDS Module Parser and glioblastoma Synthea module, are available as open source code.\footnote{\url{https://gitlab.cc-asp.fraunhofer.de/canconnect/synthea-canconnect-generator}}
In this context we also provide another Synthea module for generating patients with multiple types of cancer (without disease progressions), which is based on incidence data of the German Centre for Cancer Registry Data\footnote{\url{https://www.krebsdaten.de/Krebs/EN/Database/databasequery_step1_node.html}, (last accessed 2025/07/21)}
(Zentrum für Krebsregisterdaten, ZfKD)~\cite{meisegeier_2023_10022040}.

\section{Results}\label{sec:eval}

%An initial evaluation was conducted on the usability of the synthetic data generated by the Synthea Glioblastoma module developed in this paper.
%The generated data was compared with the input data from cancer registries.
%Various parameters were considered in the evaluation, covering different aspects of data quality.
%Specifically, demographic comparisons, tumor occurrence per population, survival time, and patient progression were analyzed.
%Using the glioblastoma Synthea module we generated about 100.000 synthetic patient records in the FHIR format.
For evaluating our glioblastoma Synthea module we generated about 100.000 synthetic patients in the FHIR format.
We removed all standard modules that come with Synthea in order to be able to evaluate our module and the synthetic data generated with it in isolation.
We evaluated how closely this synthetic dataset reproduces the characteristics of the original cancer registry data from the CanConnect dataset based on the parameters ages at diagnosis, tumor occurrence frequencies, survival times after diagnoses, and patients' pathways.

%\subsection{Demographic Data}
The gender distribution of the CanConnect dataset consists of approximately 60\% male and 40\% female individuals diagnosed with different types of glioblastoma.
The generated dataset contains around 52\% male and 48\% female individuals.
This difference is due to running Synthea with its default population model of the U.S. state of Massachusetts (cf.~Section~\ref{subsec:synthea}).
%We argue that this is not a problematic issue for our goal of generating representative patient pathways.
%However, this minor difference does not significantly impact data quality.
In the following, we do not break down the results by gender, as these differ only slightly from the results for the entire data set.
Using the default population model also impacts the age distribution of the generated patients.
However, we found that this has only a negligible influence on the properties of the synthetic data we are considering here.

\subsection{Tumor Occurrence Frequencies}
%To validate the methodology, the frequencies of occurrence of certain tumor types were compared between the CanConnect dataset and the synthetic dataset. 
%Only minor differences were observed, and all tumor types appeared at nearly the same frequency in both datasets. 
%Only minor differences were observed in the frequencies between the input and synthetic populations.
\begin{figure}
	\centering
	\includegraphics[width=1\linewidth]{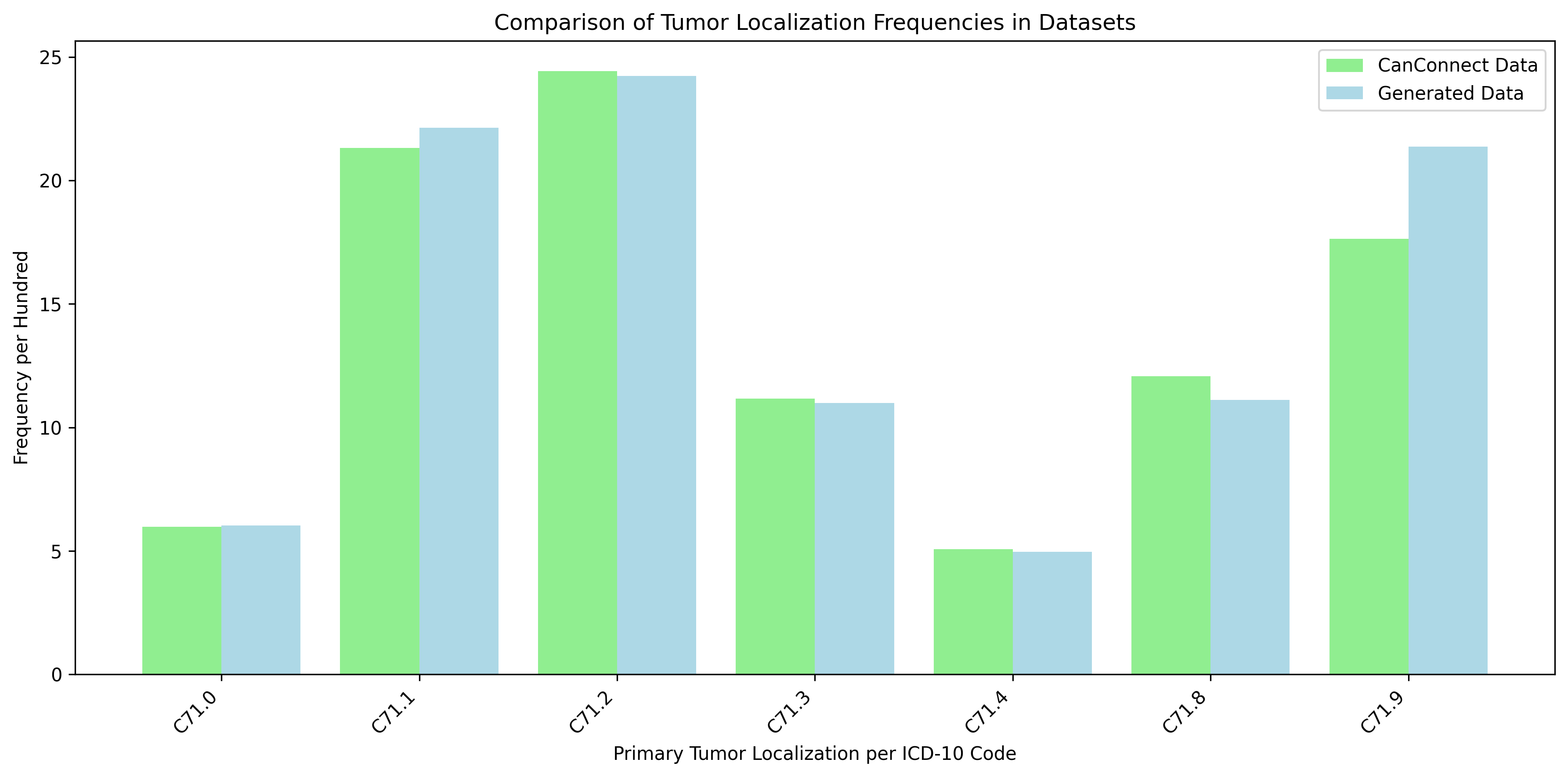}
	\caption{Comparison of the tumor frequencies in the CanConnect dataset and the generated dataset}
	\label{fig:freq}
\end{figure}
Figure~\ref{fig:freq} shows a bar chart with the occurrence frequencies of different tumor localizations (given as ICD-10 codes) for the CanConnect dataset and the synthetic dataset.
It is expected that the synthetic dataset reproduces the relative frequencies of the CanConnect dataset as the generation is based on incidences calculated from the CanConnect dataset.
Most tumor localizations appear with comparable frequencies in the synthetic dataset. % as in the CanConnect dataset.
%An even better fit would have been expected as these cases are generated based on incidences derived from the CanConnect data.
%Outliers can be observed particularly regarding the tumor types with low occurrence in the input data.
We observe the most pronounced deviation for the tumor localization $C71.9$ (malignant neoplasm: brain, unspecified), which appears 2-3 percentage points more often than in the CanConnect data.
%Also note that cases diagnosed with the ICD-10 code $C71.9$ may include more specific tumor localizations.
%, which occur frequently in the input data.
%For certain tumor types, the deviation in occurrence frequency is therefore greater than expected.
%Overall, a better fit of the tumor occurrence frequencies would have been expected as the cases are generated based on incidences derived from the CanConnect dataset.

\subsection{Ages at Diagnoses}
\begin{figure}
    \centering
    \includegraphics[width=1\linewidth]{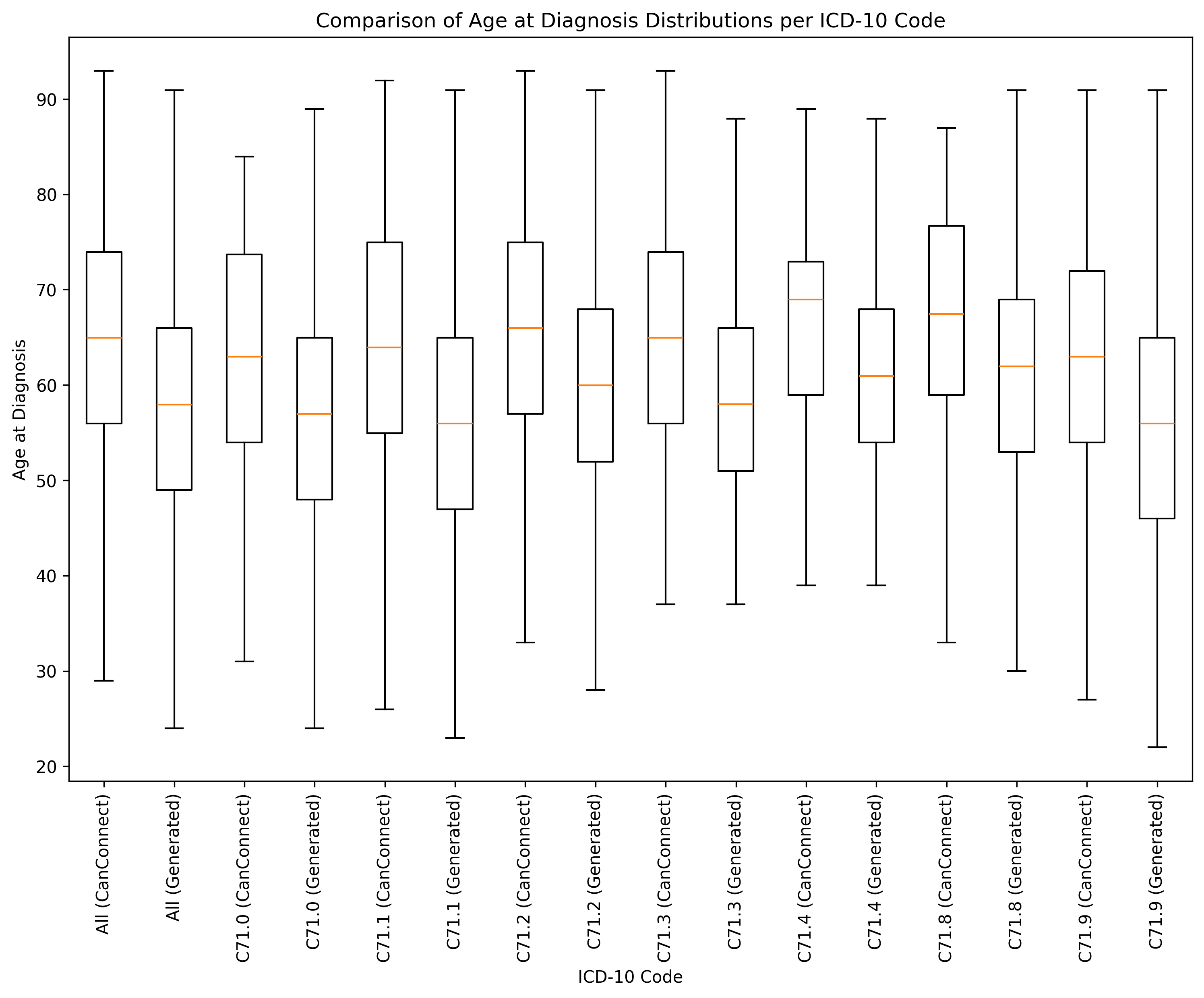}
    \caption{Boxplot of the age at diagnosis for the CanConnect dataset and the generated dataset; showing median, Q1, Q3, lower fence, and upper fence}
    \label{fig:avg_in}
\end{figure}

\begin{figure}
\begin{subfigure}[c]{0.5\textwidth}
	\includegraphics[width=0.87\textwidth]{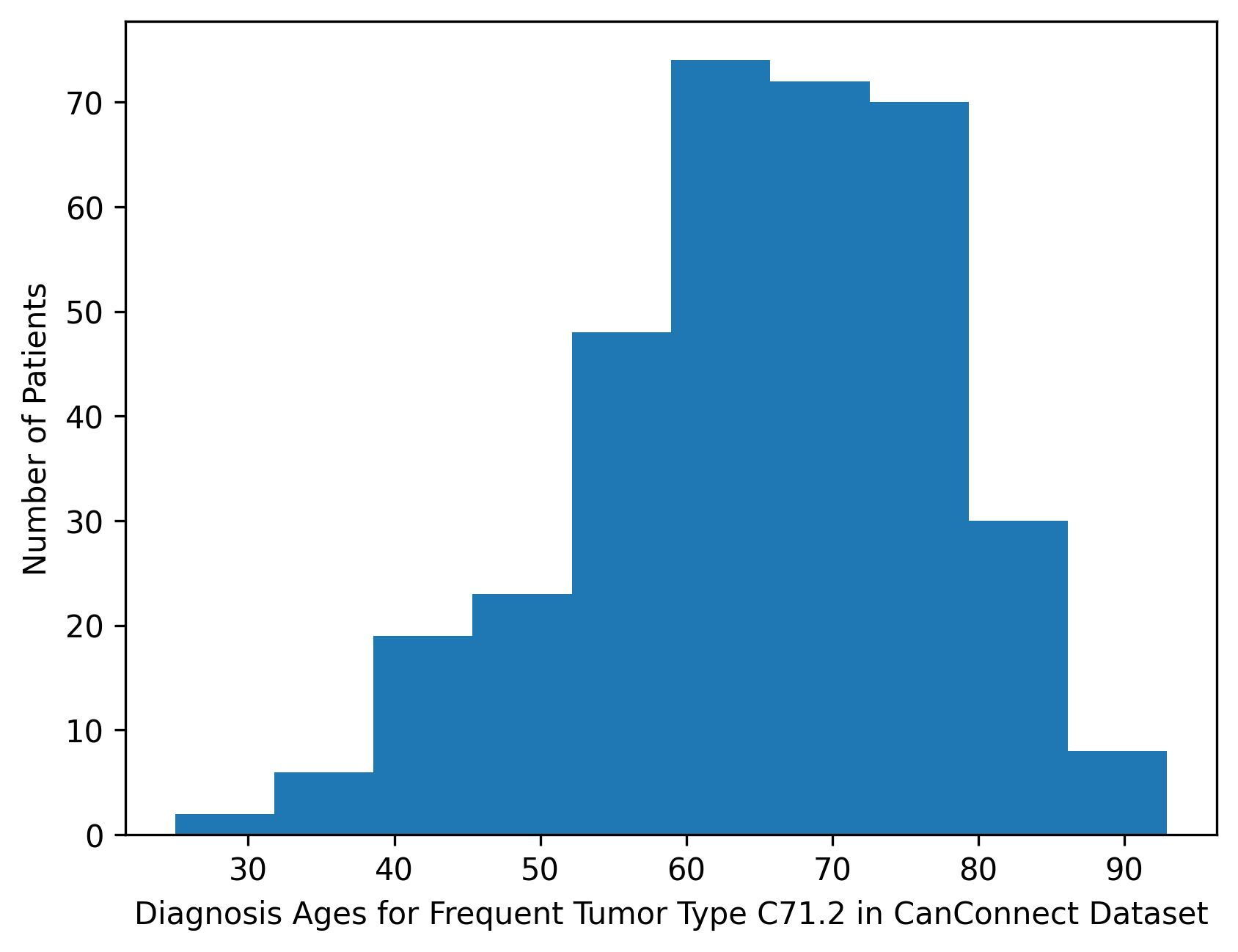}
%\subcaption{}
\end{subfigure}
\begin{subfigure}[c]{0.5\textwidth}
	\includegraphics[width=0.9\textwidth]{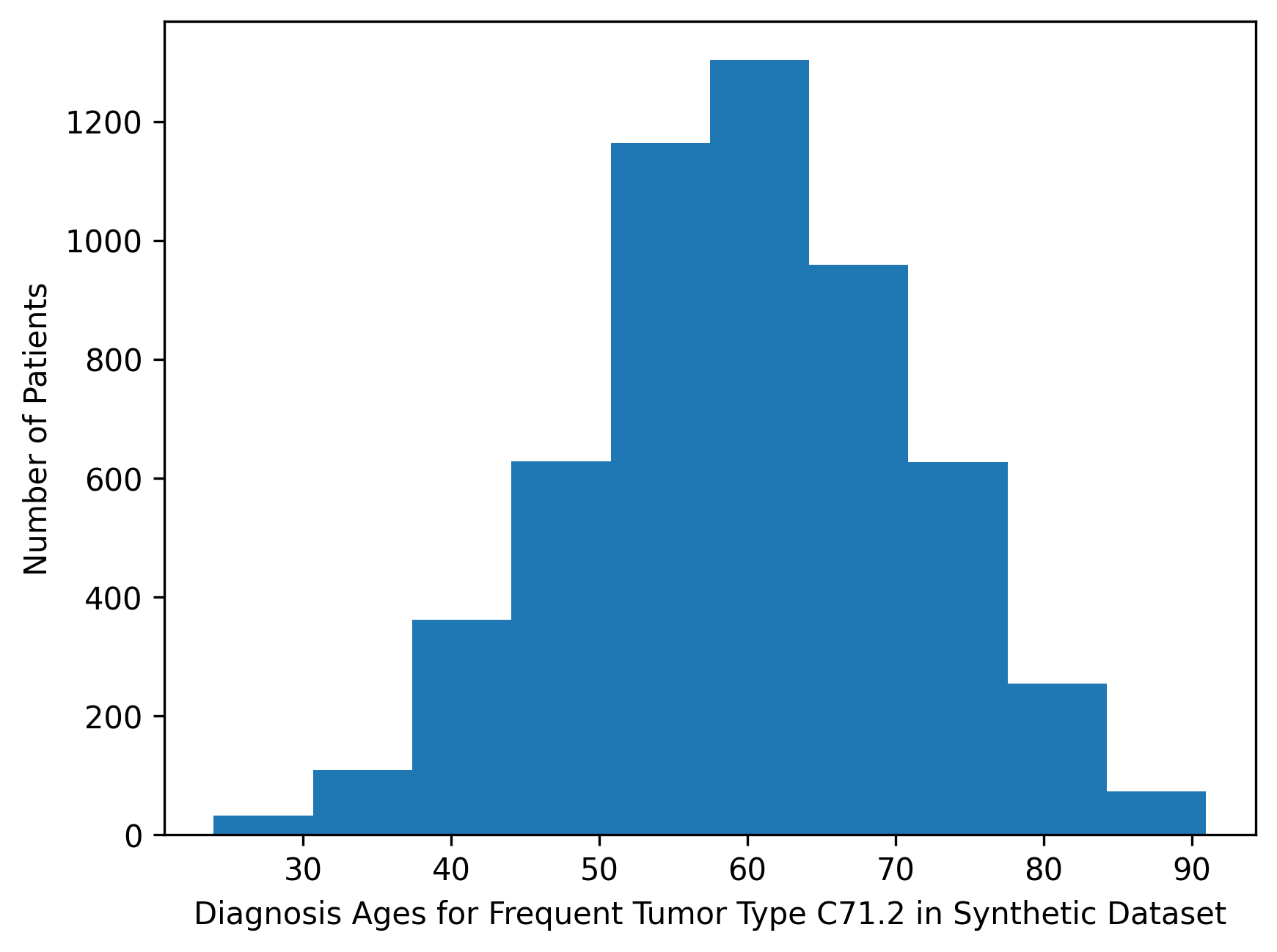}
%\subcaption{}
\end{subfigure}
    \caption{Histograms of ages at diagnosis for the most frequent tumor localization C71.2 for the CanConnect dataset and the synthetic dataset}
    \label{fig:diag-ages-dists}
\end{figure}

The average ages at diagnoses in the CanConnect dataset exhibit broad age ranges (cf.~Figure~\ref{fig:avg_in}), while most affected individuals fall within a smaller age range.
For most ICD-10 codes the CanConnect dataset also contains outliers (e.g., very young patients).
As can also be seen in Figure~\ref{fig:avg_in}, distributions for age at diagnosis for individual ICD-10 codes in the synthetic dataset are shifted downward on the Y-axis.
Using the Gaussian distribution estimated from the CanConnect dataset we obtain more diagnoses at younger ages.
This is because the Gaussian distribution cannot capture skewness.
We illustrate this in Figure~\ref{fig:diag-ages-dists} for the tumor localization $C71.2$, which is the most frequent type in the CanConnect dataset.
%This is at least partly due to the outliers in the CanConnect dataset.
Depending on the tumor localization, the average ages at diagnosis in the synthetic dataset are between 2.5\% and 12.5\% lower than in the CanConnect dataset.

\subsection{Survival Times After Diagnoses}
\label{subsec:survival-time}
\begin{figure}
    \centering
    \includegraphics[width=1\linewidth]{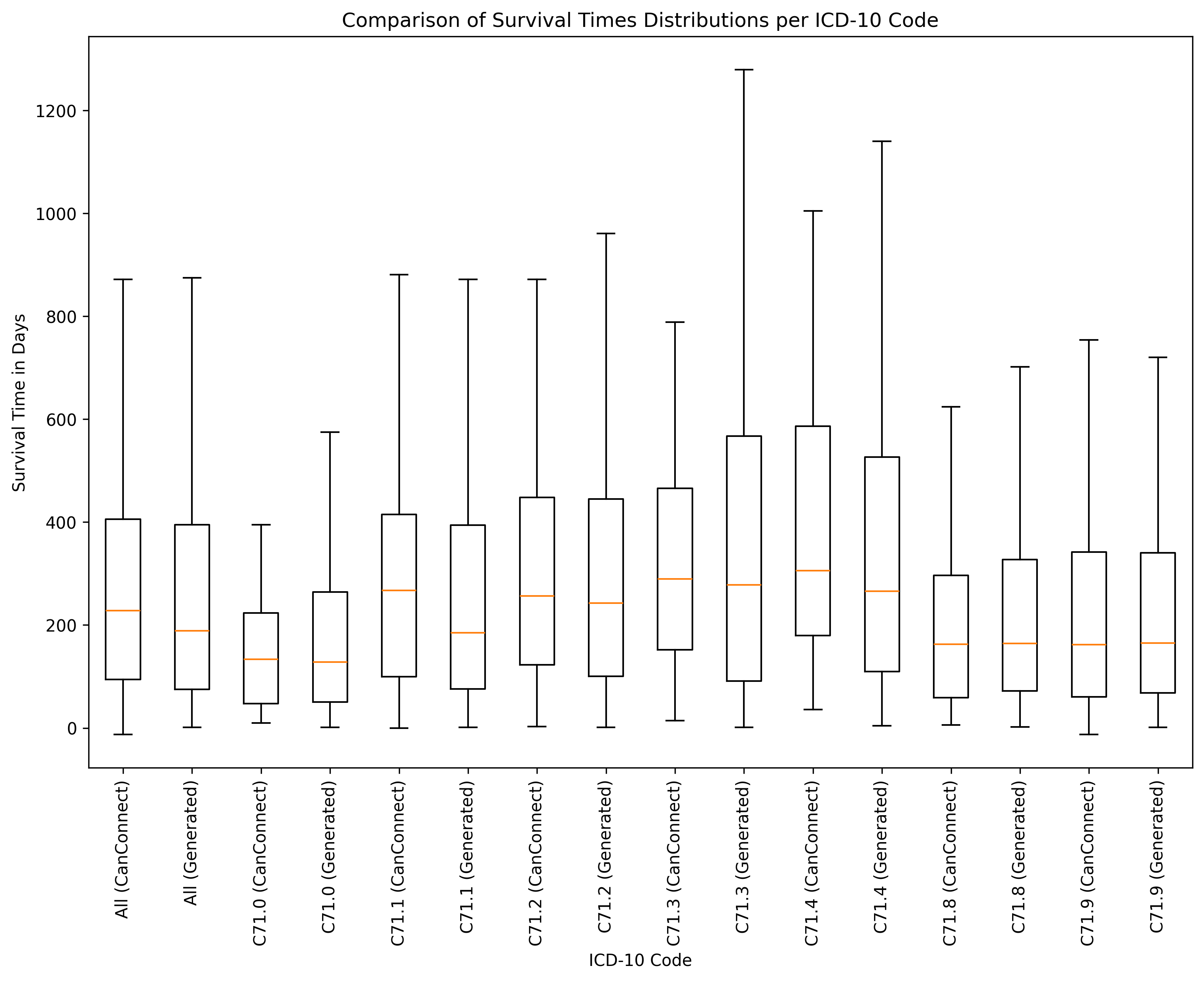}
    \caption{Boxplot of the survival time in days after diagnosis for the CanConnect dataset and the generated dataset; showing median, Q1, Q3, lower fence, and upper fence}
    \label{fig:surival_time_comparison}
\end{figure}

Figure~\ref{fig:surival_time_comparison} shows the distributions of survival times after diagnoses for the CanConnect dataset and the generated dataset for each tumor localization.
We observe that survival times distributions in the synthetic dataset are comparable to those in the CanConnect dataset.
%In the distributions of the CanConnect dataset, we see upward outliers for each tumor localization, i.\ e., patients who have survived for an exceptionally long time after the diagnosis.
%Such outliers are also created by the delay states based on exponential distributions our glioblastoma Synthea module uses to generate synthetic survival times.
%The reason we see more of these outliers for the synthetic dataset is that it contains more than an order of magnitude more cases with survival times.
Depending on tumor localization, the CanConnect dataset contains between 2 and 7.5\% outliers in survival times, while the synthetic dataset contains between 4.2 and 7.2\% of such outliers.

\subsection{Patients' Pathways}
%Finally, patient pathways were examined, specifically which OPS codes (i.e., surgical procedures) follow an ICD-coded diagnosis. 
Patient pathways were examined in terms of which surgical procedures given as OPS codes follow an ICD-10-coded diagnosis.
\begin{comment}
\begin{figure}
    \centering
    \includegraphics[width=1\linewidth]{figures/path_in_3.png}
    \caption{Pathways in the CanConnect dataset for the tumor localization C71.2}
    \label{fig:path_in}
\end{figure}
\begin{figure}
    \centering
    \includegraphics[width=1\linewidth]{figures/path_gen_3.png}
    \caption{Pathways in the generated dataset for the tumor localization C71.2}
    \label{fig:path_gen}
\end{figure}
\end{comment}
\begin{figure}
\centering
\includegraphics[width=1\linewidth]{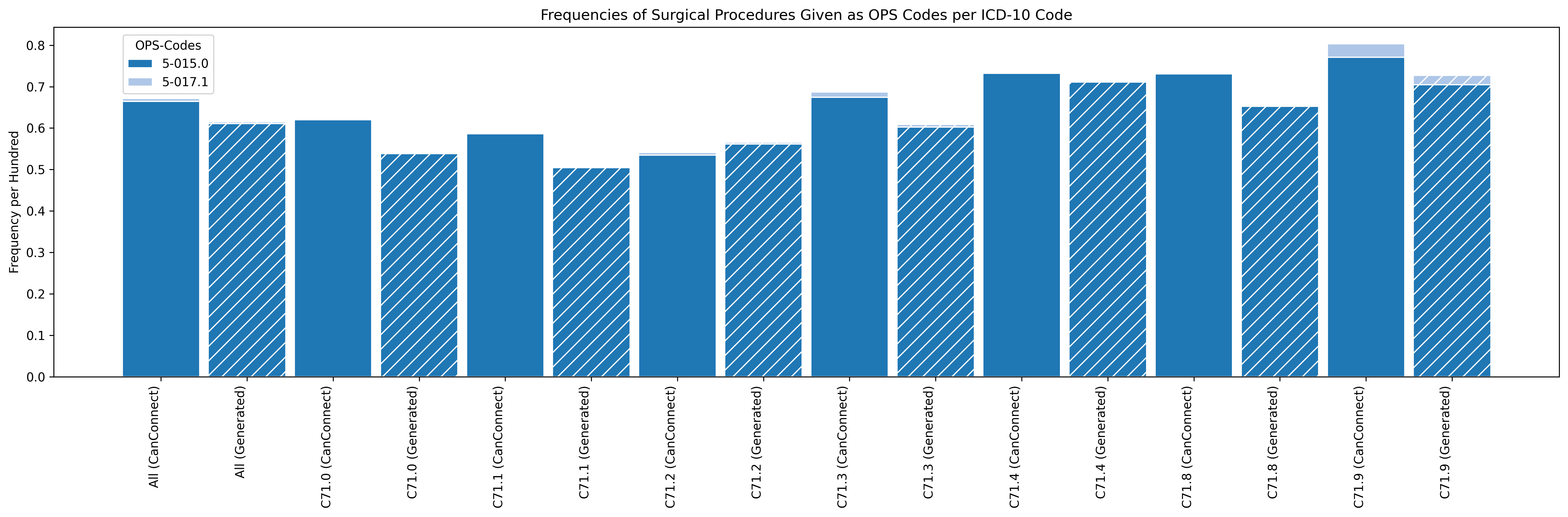}
\caption{Surgical procedures in the CanConnect dataset and in the generated dataset}
\label{fig:icd10-ops}
\end{figure}
Figure~\ref{fig:icd10-ops} shows the frequencies of surgical procedures following upon diagnoses with tumor localizations given as ICD-10 codes.
%Figures~\ref{fig:path_in} and~\ref{fig:path_gen} show exemplary Sankey diagrams of the pathways for the most frequent tumor localization $C71.2$ in the CanConnect dataset and in the generated dataset. 
%A key limitation observed is the restriction to only one possible pathway per patient in the generated data, leading to reduced variability in OPS codes.
A key limitation regarding generated patients' pathways is due to applying the restriction that a patient can have exactly one surgery or none when we generated the glioblastoma Synthea module (cf.~Section~\ref{subsec:obds-parser}). 
%An expected limitation is the reduced variety of OPS codes following upon the diagnosis of a specific tumor localization.
%This is due to applying the restriction that a patient can have exactly one surgery or none when we generated the glioblastoma Synthea module (cf.~Section~\ref{subsec:obds-parser}). 
%This is particularly the case for tumor types appearing with low frequency in the input data. 
%However, when comparing the consistency of pathways the results appear plausible.
%The occurrence frequencies of OPS codes in the generated data is slightly less close to the CanConnect dataset than expected.
Hence, we leave a more detailed examination of the generated pathways for future work, in which also the full complexity of the therapies needs to be reflected in the glioblastoma Synthea module.
%and their occurrence frequencies, the results appear plausible.
%, again influenced by tumor type frequency.

\section{Discussion}
\label{sec:discussion}
In summary, the fundamental distribution of the generated tumor data appears plausible in comparison with the CanConnect dataset, confirming that the method functions as intended.
%Relative frequencies of attributes like tumor type occurrence and OPS codes are in some cases not reproduced as similarly as expected.
%It should be noted that, as can be seen in the CanConnect data, glioblastomas are particularly rare in young people, but they do occur.
%This makes it difficult to reproduce such data synthetically using the probability distributions implemented in Synthea.
Yet, as shown for age at diagnosis, some statistical properties of the CanConnect dataset cannot be reproduced exactly using the predefined distributions implemented in Synthea.
The approach presented in this paper outlines a general way to automatically generate Synthea rules from population datasets and provides a specific example of an implementation for the oBDS format.
%The web interface enables non-experts to select specific datasets and define which disease types should be included in the generated rules to create synthetic data.
This approach has general relevance and can be extended to any datasets in a similar manner.
%The CanConnect dataset used for this publication shows a huge variety of therapy pathways.
%This is partly due to the fact that the input data was only minimally cleaned up in order to meet the goal of high automatability.
%Partly this is due to only minimal data cleansing of the data set, which, on the other hand, is sound given our goal of high automatability.
%We justify applying no further data cleansing by our goal of investigating the automatability of generating Synthea rules and synthetic data.

According to our evaluation the general concept of rule-based synthetic data generation appears feasible.
The statistics we observed in our synthetic dataset should be sufficiently realistic for many use cases, although Synthea's strength to date has been in generating realistic synthetic patient records rather than reproducing statistical properties in synthetic datasets.
%However, the achievable data quality has so far been limited by the fact that Synthea does not support any probability distributions other than the uniform distribution.
%This could be enhanced by extending Synthea to use a normal distribution or by using only the 95\% range of the data.
Our method is also limited to the reproduction of statistical properties of an input dataset, whereas disease-specific domain knowledge, for example implications of a specific biomarker or co-morbidity are not explicitly considered.
This could be improved in the future by providing the option for expert reviews of the generated rules.
Dedicated interfaces need to be developed so that domain experts can review the generated rules, whereby Synthea rulesets can quickly become very extensive as our example of glioblastoma shows.
It should also be noted that the complexity of our glioblastoma Synthea module is partly due to only minimal cleansing of the dataset, e.g., for systemic therapies we did not combine various medications with the same active ingredient.
While this is sound given our goal of high automatability of the approach, it further increases particularly the variety of therapy pathways modeled in the Synthea module.

%An examplaray application of this approach could also be to directly apply it to a networked medical device to automatically create a rule set for real data collected by the device.
%This would allow for the direct creation of a realistic research cohort on-device, which can be used without any privacy implications as no real data needs to leave the device.
%An exemplary application of this approach could be to apply it directly to a networked medical device to automatically create a rule set based on the data the device collects.
%This would enable the direct creation of a realistic research cohort on the device itself, which could be used without any privacy implications, as no real data would need to leave the device.

\section{Conclusion}\label{sec:conclusion}

%Overall, the method delivers surprisingly realistic synthethic data that shows acceptable accuracy compared to the input dataset.
The automated generation of Synthea rules based on real-world cancer registry data has great potential for a variety of applications, including training of ML models under privacy-preserving conditions and generating hypotheses, e.\ g.\ regarding new intervention strategies.
%representative data that can be used as mockups and for orientation, while reducing the patients' privacy risks associated with the need for large-scale real-world data. 
%The accuracy of some properties is acceptable enough for analysis, but users should keep in mind that some properties may be inaccurate.
%The representativeness of the generated data appears acceptable for generating hypotheses, e.\ g.\ regarding new intervention strategies.
However, medical interpretation of synthetic patient data should take into account the specific limitations associated with any currently available approach.
%Our work broadens the range of applications and specifies the strengths as well as potential limitations for the clinically relevant use case of glioblastoma, thereby contributing to a rational and evidence-based use of this powerful new technology for brain tumor research.   

\section*{Supplementary Material}
The glioblastoma Synthea module and the oBDS parser are available as open source: \url{https://gitlab.cc-asp.fraunhofer.de/canconnect/synthea-canconnect-generator}

\section*{Funding}
Research reported in this publication was supported by the German Federal Ministry of Health based on a resolution of the German Bundestag (funding codes: ZMI5-2522DAT15A, ZMI5-2522DAT15B, ZMI5-2522DAT15C, ZMI5-2522DAT15D, ZMI5-2522DAT15E).

\backmatter

%\bmhead{Supplementary information}

%If your article has accompanying supplementary file/s please state so here. 

%Authors reporting data from electrophoretic gels and blots should supply the full unprocessed scans for key as part of their Supplementary information. This may be requested by the editorial team/s if it is missing.

%Please refer to Journal-level guidance for any specific requirements.

\bmhead{Acknowledgements}

We would like to thank the Bremen Cancer Registry, the Clinical Cancer Registry Lower Saxony, and the State Cancer Registry of North-Rhine Westphalia for providing the dataset used in the evaluation.

\bibliography{sn-bibliography}% common bib file

%% BioMed_Central_Bib_Style_v1.01

\begin{thebibliography}{11}
% BibTex style file: bmc-mathphys.bst (version 2.1), 2014-07-24
\ifx \bisbn   \undefined \def \bisbn  #1{ISBN #1}\fi
\ifx \binits  \undefined \def \binits#1{#1}\fi
\ifx \bauthor  \undefined \def \bauthor#1{#1}\fi
\ifx \batitle  \undefined \def \batitle#1{#1}\fi
\ifx \bjtitle  \undefined \def \bjtitle#1{#1}\fi
\ifx \bvolume  \undefined \def \bvolume#1{\textbf{#1}}\fi
\ifx \byear  \undefined \def \byear#1{#1}\fi
\ifx \bissue  \undefined \def \bissue#1{#1}\fi
\ifx \bfpage  \undefined \def \bfpage#1{#1}\fi
\ifx \blpage  \undefined \def \blpage #1{#1}\fi
\ifx \burl  \undefined \def \burl#1{\textsf{#1}}\fi
\ifx \doiurl  \undefined \def \doiurl#1{\url{https://doi.org/#1}}\fi
\ifx \betal  \undefined \def \betal{\textit{et al.}}\fi
\ifx \binstitute  \undefined \def \binstitute#1{#1}\fi
\ifx \binstitutionaled  \undefined \def \binstitutionaled#1{#1}\fi
\ifx \bctitle  \undefined \def \bctitle#1{#1}\fi
\ifx \beditor  \undefined \def \beditor#1{#1}\fi
\ifx \bpublisher  \undefined \def \bpublisher#1{#1}\fi
\ifx \bbtitle  \undefined \def \bbtitle#1{#1}\fi
\ifx \bedition  \undefined \def \bedition#1{#1}\fi
\ifx \bseriesno  \undefined \def \bseriesno#1{#1}\fi
\ifx \blocation  \undefined \def \blocation#1{#1}\fi
\ifx \bsertitle  \undefined \def \bsertitle#1{#1}\fi
\ifx \bsnm \undefined \def \bsnm#1{#1}\fi
\ifx \bsuffix \undefined \def \bsuffix#1{#1}\fi
\ifx \bparticle \undefined \def \bparticle#1{#1}\fi
\ifx \barticle \undefined \def \barticle#1{#1}\fi
\bibcommenthead
\ifx \bconfdate \undefined \def \bconfdate #1{#1}\fi
\ifx \botherref \undefined \def \botherref #1{#1}\fi
\ifx \url \undefined \def \url#1{\textsf{#1}}\fi
\ifx \bchapter \undefined \def \bchapter#1{#1}\fi
\ifx \bbook \undefined \def \bbook#1{#1}\fi
\ifx \bcomment \undefined \def \bcomment#1{#1}\fi
\ifx \oauthor \undefined \def \oauthor#1{#1}\fi
\ifx \citeauthoryear \undefined \def \citeauthoryear#1{#1}\fi
\ifx \endbibitem  \undefined \def \endbibitem {}\fi
\ifx \bconflocation  \undefined \def \bconflocation#1{#1}\fi
\ifx \arxivurl  \undefined \def \arxivurl#1{\textsf{#1}}\fi
\csname PreBibitemsHook\endcsname

%%% 1
\bibitem[\protect\citeauthoryear{Coorevits
  et~al.}{2013}]{https://doi.org/10.1111/joim.12119}
\begin{barticle}
\bauthor{\bsnm{Coorevits}, \binits{P.}},
\bauthor{\bsnm{Sundgren}, \binits{M.}},
\bauthor{\bsnm{Klein}, \binits{G.O.}},
\bauthor{\bsnm{Bahr}, \binits{A.}},
\bauthor{\bsnm{Claerhout}, \binits{B.}},
\bauthor{\bsnm{Daniel}, \binits{C.}},
\bauthor{\bsnm{Dugas}, \binits{M.}},
\bauthor{\bsnm{Dupont}, \binits{D.}},
\bauthor{\bsnm{Schmidt}, \binits{A.}},
\bauthor{\bsnm{Singleton}, \binits{P.}},
\bauthor{\bsnm{De~Moor}, \binits{G.}},
\bauthor{\bsnm{Kalra}, \binits{D.}}:
\batitle{Electronic health records: new opportunities for clinical research}.
\bjtitle{Journal of Internal Medicine}
\bvolume{274}(\bissue{6}),
\bfpage{547}--\blpage{560}
(\byear{2013})
\doiurl{10.1111/joim.12119}
\end{barticle}
\endbibitem

%%% 2
\bibitem[\protect\citeauthoryear{Sweeney}{2002}]{Sweeney2002}
\begin{barticle}
\bauthor{\bsnm{Sweeney}, \binits{L.}}:
\batitle{$k$-anonymity: A model for protecting privacy}.
\bjtitle{International Journal of Uncertainty, Fuzziness and Knowledge-Based
  Systems}
\bvolume{10}(\bissue{05}),
\bfpage{557}--\blpage{570}
(\byear{2002})
\doiurl{10.1142/s0218488502001648}
\end{barticle}
\endbibitem

%%% 3
\bibitem[\protect\citeauthoryear{Walonoski et~al.}{2017}]{syntheaRef}
\begin{botherref}
\oauthor{\bsnm{Walonoski}, \binits{J.}},
\oauthor{\bsnm{Kramer}, \binits{M.}},
\oauthor{\bsnm{Nichols}, \binits{J.}},
\oauthor{\bsnm{Quina}, \binits{A.}},
\oauthor{\bsnm{Moesel}, \binits{C.}},
\oauthor{\bsnm{Hall}, \binits{D.}},
\oauthor{\bsnm{Duffett}, \binits{C.}},
\oauthor{\bsnm{Dube}, \binits{K.}},
\oauthor{\bsnm{Gallagher}, \binits{T.}},
\oauthor{\bsnm{McLachlan}, \binits{S.}}:
{Synthea: An approach, method, and software mechanism for generating synthetic
  patients and the synthetic electronic health care record}.
Journal of the American Medical Informatics Association
\textbf{25}(3)
(2017)
\doiurl{10.1093/jamia/ocx079}
\end{botherref}
\endbibitem

%%% 4
\bibitem[\protect\citeauthoryear{Braunstein}{2018}]{fhirStandard}
\begin{bbook}
\bauthor{\bsnm{Braunstein}, \binits{M.L.}}:
\bbtitle{Health Informatics on FHIR: How HL7's New API Is Transforming
  Healthcare}.
\bpublisher{Springer},
\blocation{Cham}
(\byear{2018})
\end{bbook}
\endbibitem

%%% 5
\bibitem[\protect\citeauthoryear{Stadler et~al.}{2020}]{Stadler2021}
\begin{botherref}
\oauthor{\bsnm{Stadler}, \binits{T.}},
\oauthor{\bsnm{Oprisanu}, \binits{B.}},
\oauthor{\bsnm{Troncoso}, \binits{C.}}:
Synthetic data -- anonymisation groundhog day
(2020)
{\href{https://arxiv.org/abs/2011.07018}{{2011.07018}}}
\end{botherref}
\endbibitem

%%% 6
\bibitem[\protect\citeauthoryear{Buczak et~al.}{2010}]{Buczak2010}
\begin{botherref}
\oauthor{\bsnm{Buczak}, \binits{A.L.}},
\oauthor{\bsnm{Babin}, \binits{S.}},
\oauthor{\bsnm{Moniz}, \binits{L.}}:
Data-driven approach for creating synthetic electronic medical records.
BMC Medical Informatics and Decision Making
\textbf{10}(1)
(2010)
\doiurl{10.1186/1472-6947-10-59}
\end{botherref}
\endbibitem

%%% 7
\bibitem[\protect\citeauthoryear{Al~Aziz et~al.}{2021}]{10.1145/3469035}
\begin{botherref}
\oauthor{\bsnm{Al~Aziz}, \binits{M.M.}},
\oauthor{\bsnm{Ahmed}, \binits{T.}},
\oauthor{\bsnm{Faequa}, \binits{T.}},
\oauthor{\bsnm{Jiang}, \binits{X.}},
\oauthor{\bsnm{Yao}, \binits{Y.}},
\oauthor{\bsnm{Mohammed}, \binits{N.}}:
Differentially private medical texts generation using generative neural
  networks.
ACM Trans. Comput. Healthcare
\textbf{3}(1)
(2021)
\doiurl{10.1145/3469035}
\end{botherref}
\endbibitem

%%% 8
\bibitem[\protect\citeauthoryear{Hernandez et~al.}{2022}]{HERNANDEZ202228}
\begin{barticle}
\bauthor{\bsnm{Hernandez}, \binits{M.}},
\bauthor{\bsnm{Epelde}, \binits{G.}},
\bauthor{\bsnm{Alberdi}, \binits{A.}},
\bauthor{\bsnm{Cilla}, \binits{R.}},
\bauthor{\bsnm{Rankin}, \binits{D.}}:
\batitle{Synthetic data generation for tabular health records: A systematic
  review}.
\bjtitle{Neurocomputing}
\bvolume{493},
\bfpage{28}--\blpage{45}
(\byear{2022})
\doiurl{10.1016/j.neucom.2022.04.053}
\end{barticle}
\endbibitem

%%% 9
\bibitem[\protect\citeauthoryear{Walonoski et~al.}{2020}]{WALONOSKI2020100007}
\begin{barticle}
\bauthor{\bsnm{Walonoski}, \binits{J.}},
\bauthor{\bsnm{Klaus}, \binits{S.}},
\bauthor{\bsnm{Granger}, \binits{E.}},
\bauthor{\bsnm{Hall}, \binits{D.}},
\bauthor{\bsnm{Gregorowicz}, \binits{A.}},
\bauthor{\bsnm{Neyarapally}, \binits{G.}},
\bauthor{\bsnm{Watson}, \binits{A.}},
\bauthor{\bsnm{Eastman}, \binits{J.}}:
\batitle{Synthea™ novel coronavirus (covid-19) model and synthetic data set}.
\bjtitle{Intelligence-Based Medicine}
\bvolume{1-2},
\bfpage{100007}
(\byear{2020})
\doiurl{10.1016/j.ibmed.2020.100007}
\end{barticle}
\endbibitem

%%% 10
\bibitem[\protect\citeauthoryear{Walonoski et~al.}{2018}]{walonoski2018synthea}
\begin{barticle}
\bauthor{\bsnm{Walonoski}, \binits{J.}},
\bauthor{\bsnm{Kramer}, \binits{M.}},
\bauthor{\bsnm{Nichols}, \binits{J.}},
\bauthor{\bsnm{Quina}, \binits{A.}},
\bauthor{\bsnm{Moesel}, \binits{C.}},
\bauthor{\bsnm{Hall}, \binits{D.}},
\bauthor{\bsnm{Duffett}, \binits{C.}},
\bauthor{\bsnm{Dube}, \binits{K.}},
\bauthor{\bsnm{Gallagher}, \binits{T.}},
\bauthor{\bsnm{McLachlan}, \binits{S.}}:
\batitle{Synthea: An approach, method, and software mechanism for generating
  synthetic patients and the synthetic electronic health care record}.
\bjtitle{Journal of the American Medical Informatics Association}
\bvolume{25}(\bissue{3}),
\bfpage{230}--\blpage{238}
(\byear{2018})
\end{barticle}
\endbibitem

%%% 11
\bibitem[\protect\citeauthoryear{Meisegeier
  et~al.}{2023}]{meisegeier_2023_10022040}
\begin{botherref}
\oauthor{\bsnm{Meisegeier}, \binits{S.}},
\oauthor{\bsnm{Imhoff}, \binits{M.}},
\oauthor{\bsnm{Berg}, \binits{K.}},
\oauthor{\bsnm{Kraywinkel}, \binits{K.}}:
Bundesweiter Klinischer Krebsregisterdatensatz - Datenschema und
  Klassifikationen.
\doiurl{10.5281/zenodo.10022040}
\end{botherref}
\endbibitem

\end{thebibliography}
%% if required, the content of .bbl file can be included here once bbl is generated
%%\input sn-article.bbl

\end{document}